\def\year{2018}\relax
\documentclass[letterpaper]{article} %
\usepackage{aaai18}  %
\usepackage{times}  %
\usepackage{helvet}  %
\usepackage{courier}  %
\usepackage{url}  %
\usepackage{graphicx}  
\usepackage[utf8]{inputenc}
\usepackage[table]{xcolor}
\usepackage{epsfig}
\usepackage{pgfplotstable}
\usepackage{pgfplots}
\pgfplotsset{compat=1.5}
\usepackage{amsmath}
\usepackage{amssymb}
\usepackage{bbm}
\usepackage{float}
\usepackage{hyperref}
\usepackage{microtype}
\usepackage{tikz}
\usetikzlibrary{calc,shapes,arrows,arrows.meta, positioning,shapes.misc,decorations.markings,shapes.geometric}
\usepackage{booktabs}
\usepackage{multirow}
\usepackage{adjustbox}
\usepackage{verbatim}
\usepackage[T1]{fontenc}
\usepackage[title]{appendix}
\usepackage[font={small,singlespacing},labelfont=it,labelsep=period]{caption}
\usepackage{adjustbox} %
\usepackage{graphicx}
\usepackage{caption}
\usepackage{subcaption}
\usepackage[colorinlistoftodos]{todonotes}
\usepackage{wrapfig}
\usepackage{pdfpages}
\usepackage{nicefrac}
\usepackage[]{placeins}%

\newcommand{\citet}[1]
  {\citeauthor{#1} ~\shortcite{#1}}
  \newcommand{\citep}{\cite}
  
\newcommand\textcite[1]{\citet{#1}}

\usepackage[autostyle, english=american]{csquotes}
\MakeOuterQuote{"}

\makeatletter
\pgfplotsset{
    every axis x label/.append style={
        alias=current axis xlabel
    },
    legend pos/outer south/.style={
        /pgfplots/legend style={
            at={%
                (%
                \@ifundefined{pgf@sh@ns@current axis xlabel}%
                {xticklabel cs:0.5}%
                {current axis xlabel.south}%
                )%
            },
            anchor=north
        }
    }
}
\makeatother

\newcolumntype{t}{>{\ttfamily}l}
\newcolumntype{T}{>{\ttfamily}c}

\newcolumntype{$}{>{\global\let\currentrowstyle\relax}}
\newcolumntype{^}{>{\currentrowstyle}}

\renewcommand{\todo}[2][]{\tikzexternaldisable\@todo[#1]{#2}\tikzexternalenable}

\begin{document}

\everypar{\looseness=-1} %
\linepenalty=100 %

\title{Malware Detection by Eating a Whole EXE%
}

  \author{Edward Raff\textsuperscript{1,3,4},
   Jon Barker\textsuperscript{2},
   Jared Sylvester\textsuperscript{1,3},
   Robert Brandon\textsuperscript{1,3,4}\\
   {\bf \Large Bryan Catanzaro\textsuperscript{2},
   Charles Nicholas\textsuperscript{4}}\\
  \textsuperscript{1}{Laboratory for Physical Sciences},
  \textsuperscript{2}{NVIDIA},
  \textsuperscript{3}{Booz Allen Hamilton},
  \textsuperscript{4}{University of Maryland, Baltimore County}\\
  \{edraff,jared,rbrandon\}@lps.umd.edu, \{jbarker,bcatanzaro\}@nvidia.com, nicholas@umbc.edu
  }

\maketitle

\begin{abstract}
In this work we introduce malware detection from raw byte sequences as a fruitful research area to the larger machine learning community. Building a neural network for such a problem presents a number of interesting challenges that have not occurred in 
tasks such as image processing or NLP. In particular, we note that detection from raw bytes presents a sequence problem with over two million time steps and a problem where batch normalization appear to hinder the learning process. We present our initial work in building a solution to tackle this problem, which has linear complexity dependence on the sequence length, and allows for interpretable sub-regions of the binary to be identified. In doing so we will discuss the many challenges in building a neural network to process data at this scale, and the methods we used to work around them.  

\end{abstract}

\section{Introduction}

The detection of malicious software (malware) is an important problem in cyber security, especially as more of society becomes dependent on computing systems. Already, single incidences of malware can cause millions of dollars in 
damages~\cite{Anderson2013}. 
Anti-virus products provide some protection against malware, but are growing increasingly ineffective for the problem. Current anti-virus technologies use a signature-based approach, where a signature is a set of manually crafted rules %
in an attempt to identify a small family of malware. These rules are generally specific, and cannot usually recognize new malware even if it uses the same functionality. This approach is insufficient as most environments will have unique binaries that will have never been seen before~\cite{Li2017} and millions of new malware samples are found every day. The limitations of signatures have been recognized by the anti-virus providers and industry experts for many years~\cite{tagkey2014iv}. The need to develop techniques that generalize to new malware would make the task of malware detection a seemingly perfect fit for machine learning, though there exist significant challenges. 

To build a malware detection system, we must first determine a feature set to use. One intuitive choice is to use features obtained by monitoring program execution (APIs called, instructions executed, IP addresses accessed, etc.). This is referred to as dynamic analysis. While intuitively appealing, there are many issues with dynamic analysis in practice. To conduct dynamic analysis, malware must be run inside a specially instrumented environment, such as a customized Virtual Machine (VM), which introduces high computational requirements. Furthermore, in some cases it is possible for malware to detect when it is being analyzed. When the malware detects an attempt to analyze it, the malware can  alter its behavior, allowing it to avoid discovery~\cite{Raffetseder:2007:DSE:2396231.2396233,Garfinkel:2007:CTV:1361397.1361403,Carpenter:2007:HVA:1271924.1272057}. Even when malware does not exhibit this behavior, the analysis environment may not reflect the target environment of the malware, creating a discrepancy between the training data collected and real life environments~\cite{Rossow2012}. While a dynamic analysis component is likely to be an important component for a long term solution, we avoid it at this time due to its added complexity. 

We instead take a static analysis approach, where we look at information from the binary program that can be obtained without running it. In particular, we look at the raw bytes of the file itself, and build a neural network to determine maliciousness. 
Neural nets have excelled in learning features from raw inputs for image~\cite{szegedy2015going}, signal~\cite{graves2013speech}, and text~\cite{zhang2015text} problems.
Replicating this success in the malware domain may help to simplify the tools used for detecting malware and improve accuracy. Because malware may frequently exploit bugs and ignore format specifications, parsing malicious files and using features that require domain knowledge can require significant and nontrivial effort. Since malware is written by a real live adversary, such code will also require maintenance and improvement to adjust to changing behavior of the malware authors.
Since we desire to learn a system from raw byte inputs, from which higher level representations will be constructed, we choose to use a neural network based approach. However, there exist a number of %
challenges and differences for this domain that  have not been encountered in other tasks. These challenges make research in malware detection intrinsically interesting and relevant from a machine learning perspective beyond merely introducing these techniques to a novel domain. For Microsoft Windows Portable Executable (PE) malware, these challenges include but are not limited to:

\begin{enumerate}
\item The bytes in malware can have multiple modalities of information. The meaning of any particular byte is context sensitive, and could be encoding human-readable text (e.g., function names from the import table), binary code, arbitrary objects such as images (from the resource/data sections of a binary), and more. 
\item The content of a binary exhibits multiple types of spatial correlation. Code instructions in a function are intrinsically correlated spatially, but this correlation has discontinuities from function calls and jump commands. Further, the contents at a function level can be arbitrarily re-arranged if addresses are properly corrected. 
\item Treating each byte as a unit in a sequence, we are dealing with a sequence classification problem on the order of \textit{two million time steps}. To our knowledge, this far exceeds the length of input to any previous neural network based sequence classifier.
\item Our problem has multiple levels of concept drift over time. The applications, build tools, and libraries developers use will naturally be updated, and alternatives will fall in and out of favor. This alone causes concept drift. But malware is written by a real-life adversary, and is often intentionally adjusted to avoid detection. 
\end{enumerate}

Our contributions in this work are the development of the first, to our knowledge, network architecture that can successfully process a raw byte sequence of over two million steps. Others have attempted this task, but failed to outperform simpler baselines or successfully process the entire file~\cite{Anderson2017}, in part because the techniques developed for signal and image processing do not always transfer to this new domain. We identify the challenges involved in making a network detect malware from raw bytes, and the initial methods one can employ to successfully train such a model. We show that this model learns a wider breadth of information types compared to previous domain-knowledge free approaches to malware detection. Our work also highlights a failure case for batch-normalization, which initially rendered our model unable to learn.

\section{Related work}

There are two primary themes of past work: the application of neural networks to ever longer sequences, and
the application of neural networks to malware detection. 
The use of Recurrent Neural Networks (RNNs) has been historically prevalent in any work involving sequences, but the processing of raw bytes far exceeds the scale attempted in previous work by orders of magnitude. For malware detection, 
all of these previous applications use a significant amount of domain knowledge for feature extraction. In contrast, our goal is to minimize the use of such domain knowledge, and explore how much of the problem can be solved without specifying any such information. 

\subsection{Neural Networks for Long Sequences}

Little work has been done on the scale of sequence classification explored in this work. The closest in terms of pure sequence length is WaveNet~\cite{wavenet}. WaveNet attempts to advance the state-of-the-art in generative audio by ignoring previous feature engineering, and instead using the raw bytes of the audio as the input feature and target. This results in a sequence problem with 16,000 time steps per second of audio.
Wide receptive fields for this task (4,800 steps) were obtained through the use of dilated convolutions~\cite{Yu2016} and by training a very deep architecture. Ultimately, their work is still on the order of two magnitudes smaller in sequence length compared to our malware detection problem. 

The use of dilated convolutions to handle sequence length has become a common trend, as for example in the ByteNet model for machine translation~\cite{Kalchbrenner2016a}. While translation can result in relatively long sequences, their sequence length is smaller than WaveNet's audio generation. While we did explore dilated convolutions in this work, we did not find them to perform any better or worse than standard convolutions for our problem. We suspect this is due the different nature of locality in binaries, that the values in the "holes" of the dilation are easier to assume or interpolate for spatially consistent domains like image classification, but are not obviously interpolated for binary content. 

We note another trend when working with long sequences: the use of RNNs that operate at different frequencies. \textcite{Mehri2017} used such an architecture for audio classification, but exploited the generative nature of the task to train on sub-sequences of only 512 time steps. Other works that have made use of RNNs operating at multiple frequencies have similarly worked on sequences that do not exceed thousands of time steps~\cite{koutnik2014clockwork,NIPS2016_6310}.

In addition to the difficulties in dealing with the unusually long sequences that we confront, we must also contend with a lack of information flow.
When making a benign/malicious classification of a binary we obtain only one error signal, which must be used to inform decisions regarding all 2 million time steps.
In contrast, neural translation models and autoregressive models such as WaveNet are attempting to predict not an overall classification, but the next word or byte. This provides them with frequent label information at each time step, resulting in a near 1:1 mapping between input size and labels from which to propagate errors. Such frequent gradient information is not available for our problem, increasing the learning challenge even before considering sequence length.

\subsection{Neural Networks for Malware Detection}

There has been little work thus far in applying neural networks to malware detection, and no current work we are aware of that attempts to do so from the raw bytes of the entire binary.
It has recently been 
demonstrated that fully connected and recurrent networks are able to learn the malware identification problem when trained on just 300 bytes from the PE-header of each file~\cite{raff2017peheader}. Based on the positive results obtained, the current work extends those results by training networks on entire, several million byte long executables, and encounters a wide breadth of potential byte content.

The work of \textcite{Saxe2015a} is closest to ours at a feature level, as it uses a histogram of byte entropy values for features. This is in addition to a histogram of ASCII string lengths, PE imports, and other meta-data that can be obtained via static analysis. This approach produces some small level of information from the whole file, but discards most information about the actual content of the binary in the process, as it creates a fixed length feature vector to use as input to the network. 

Most recent work in the application of deep learning to malware detection has used features extracted via dynamic analysis, where the binary is run in a virtualized environment to obtain information about its execution. \textcite{Kolosnjaji2016} tackled the related problem of malware family classification (i.e., which family does a particular malicious file belong to?) using a combination of convolutions followed by LSTMs to process the sequence of API calls a malware file generated under dynamic analysis. This was after down-selecting to just 60 kernel API calls to track. 

\textcite{Huang:2016:MMN:2976956.2976984} performed manual feature engineering of API calls into 114 higher-level concepts, and combined these API events with input arguments to the original function calls as well as tri-grams. Rather than just predict maliciousness, they performed malware detection and family classification with the same model (i.e., weights shared between two tasks). This approach improved the performance of the model on both tasks, and would be compatible with our design in this work. 

These prior works in malware detection tend to use significant manual feature engineering, which requires a significant if not rare level of domain expertise. Those using dynamic analysis often rely on sophisticated non-public emulation environments to mitigate the challenges with dynamic analysis, which significantly increases the effort to reproduce work. Our proposed approach eliminates this domain knowledge-specific code and feature processing, reducing the amount of specialized code and reducing the barrier to reproduction and extension. %

We note one unfortunate aspect of much of the previous work in malware detection, including some of our own, namely, the use of data that is not available to the public, for various reasons.  %
Data that can be readily obtained by the public is often not of a sufficient quality to be usable in practice, as we will discuss in \autoref{sec:training_data}. This also means we cannot meaningfully compare accuracy numbers across works, as different datasets are used with different labeling procedures. 

\section{Training data} \label{sec:training_data}

For this work we use the same training and testing data as in \textcite{raff_ngram_2016}. Specifically, we use the Group B training data, and Group A \& B testing data. Group B data was provided by an anti-virus industry partner, where both the benign and malicious programs are meant to be representative of files seen on real machines. The Group B training set consists of 400,000 files split evenly between benign and malicious classes. The testing set has 77,349 files, of which 40,000 are malicious and the remainder are benign. 

The Group A data was collected in the same manner as most work in the malware identification literature
~\cite{Schultz2001,Kolter:2006:LDC:1248547.1248646}
, which is available to the public. The benign data (or "goodware") comes from a clean installation of Microsoft Windows, with some commonly installed applications (e.g., Firefox, Flash, etc) and the malware comes from the VirusShare corpus~\cite{VirusShare}. The Group A test set contains 43,967 malicious and 21,854 benign testing files. 

It was found that training on the Group A-style data results in severe overfitting~\cite{raff_ngram_2016}, learning to recognize "from Microsoft" instead of "benign", which does not generalize to new data. That is to say, a model trained on Group A doesn't generalize to Group B, but a model trained on Group B does generalize to Group A. For this reason we only perform our experiments with the Group B training data, and test on both groups. Testing in this manner allows us to better quantify generalization ability, as the data are from different sources. This minimizes shared biases, and gives us a potential upper and lower-bound on expected accuracy. 

\begin{figure}
  \begin{center}
  \begin{adjustbox}{width=\columnwidth}
    \begin{tikzpicture}[auto]

      \node (byte) [rectangle, draw=black] {Raw Byte};

      \node (embd) [rectangle, rounded corners, draw=black, right of=byte, xshift=+1.4cm] {Embedding};

      \node (conv_gate) [rectangle, rounded corners, draw=black, below right of=embd, xshift=+0.2cm, yshift=-0.2cm] {1D Conv};

      \node (conv_base) [rectangle, rounded corners, draw=black, above right of=embd, xshift=+0.2cm, yshift=+0.2cm] {1D Conv};

      \node (sigma) [circle, rounded corners, draw=black, right of=conv_gate, minimum height=0.8cm, xshift=+0.6cm] {$\sigma$};

      \node (mult) [circle, rounded corners, draw=black, right of=conv_base, minimum height=0.8cm, xshift=+0.6cm] {$\otimes$};

      \node (max_pool) [rectangle, rounded corners, draw=black, right of=mult, xshift=+1.6cm] {Temporal Max-Pooling};

      \node (flc) [rectangle, rounded corners, draw=black, below of=max_pool, yshift=+0.1cm] {Fully Connected};

      \node (out) [rectangle, rounded corners, draw=black, below of=flc, yshift=+0.1cm] {Softmax};

      \draw [thick,->,>=stealth] (byte) --  node[anchor=east] {} (embd);

      \draw [thick,->,>=stealth] (embd) --  (conv_gate);
      \draw [thick,->,>=stealth] (embd) --  (conv_base);

      \draw [thick,->,>=stealth] (conv_gate) --  (sigma);
      \draw [thick,->,>=stealth] (conv_base) --  (mult);

      \draw [thick,->,>=stealth] (sigma) --  (mult);

      \draw [thick,->,>=stealth] (mult) --  (max_pool);

      \draw [thick,->,>=stealth] (max_pool) --  (flc);

      \draw [thick,->,>=stealth] (flc) --  (out);

    \end{tikzpicture}
    \end{adjustbox}
  \end{center}
  \caption{Architecture diagram of MalConv model.}
  \label{fig:model_block_diagram}
\end{figure}
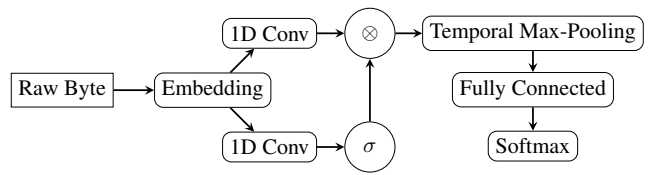

We use both group's test sets, as this allows us to better judge the generalization ability of the models. 
Group B's test performance is important, as it is supposed to represent data in the wild, but may have a shared common bias due to how Group B data was collected. 
Testing on the Group A data, which is collected in a different manner, then is a stronger test of generalization as the data has fewer common biases with Group B. Because of this, we consider Group A's test performance more interesting than Group B's. We also want our model to have similar performance on both test sets, which would indicate the features learned are widely useful.

In addition, reaching out to the original company, we have obtained a larger  training corpus of 2,011,786 binaries, with 1,000,020 benign and 1,011,766 malicious. We use this larger dataset to show that our new MalConv architecture continues to improve with increased training data, while the byte n-gram approach appears to have plateaued in terms of performance.

\section{Model Architecture}

When designing our model three features were desired: 1) the ability to scale well with sequence length, 2) the ability to consider both local and global context while examining an entire file, and 3) an explanatory ability to aid analysis of flagged malware. A block diagram of this model, which we refer to as MalConv, is given in \autoref{fig:model_block_diagram}, and a more detailed diagram is in the supplemental material. 

Our architectural choices were influenced in large part by the need to address the high amount of positional variation present in executable files. At a high level, the contents of a PE binary can be rearranged in almost any arbitrary ordering. The only fixed constant is the MS-DOS header, which ends with a pointer to the beginning of the PE-Header. The PE-Header can then be anywhere, and parts of it can be located throughout the file. The PE-Header itself contains pointers to all other contents of the binary (code, resources, etc). This allows a macro-reorganization of the byte contents without ever changing the meaning. Similarly, even within the code sections of a binary, the definition of functions can be re-ordered so long as address of sets used in the code are correctly adjusted. This is another level of spatial restructuring that can occur. This macro-level reordering represents one of many types of spatial properties within a binary, but we consider it to be the most important to tackle. Spatial discontinuities at a function level will remain difficult, but are not insurmountable for the model to learn around. Correlations across large ranges will likely be missed; we hope to capture that information in future work.

To best capture such high level location invariance, we choose to use a convolution network architecture. Combining the convolutional activations with a global max-pooling before going to fully connected layers allows our model to produce its activation regardless of the location of the detected features. Rather than perform convolutions on the raw byte values (i.e., using a scaled version of a byte's value from 0 to 255), we use an embedding layer to map each byte to a fixed length (but learned) feature vector. We avoid the raw byte value as it implies an interpretation that certain byte values are intrinsically "closer" to each-other than other byte values, which we know a priori to be false, as byte value meaning is dependent on context. Training the embedding jointly with the convolution allows even our shallow network to activate for a wider breadth of input patterns. This also gives it a degree of robustness in the face of minor alterations in byte values. Prior work using byte n-grams lack this quality, as they are dependent on exact byte matches~\cite{Kolter:2006:LDC:1248547.1248646,raff_ngram_2016}.

We note a number of difficult design choices that had to be made in developing a neural network architecture for such long input sequences. One of the primary limitations in practice was GPU memory consumption in the first convolution layer. Regardless of convolution size, storing the activations after the first convolution for forward propagation can easily lead to out-of-memory errors during back-propagation. We chose to use large convolutional filters and strides to control the memory used by activations in these early layers. 

Attempts to build deep architectures on such long sequences requires aggressive pooling between layers for our data, which results in lopsided memory use. This makes model parallelism in frameworks like Tensorflow %
difficult to achieve. 
Instead we chose to create a shallow architecture with a large filter width of 500 bytes combined with an aggressive stride of 500. This allowed us to better balance computational workload in a data-parallel manner using PyTorch~\cite{pytorchGithub}. Our convolutional architecture uses the gated convolution approach following \textcite{dauphin2016language}, with 128 filters.

\subsubsection{Regularization}

A consistent result across tested architectures is a propensity for overfitting. This is not surprising given the large size of our input feature space (2 million time steps) from which we must learn the benign/malicious classification based off a single loss. In particular we note the %
difficulty in generalizing from both the Group B training data to the Group B testing data, as well as the Group B training data to the Group A test data. In development we found the DeCov regularization~\cite{Cogswell2016} to be most helpful, which penalizes correlation between the hidden state activations at the penultimate layer. \looseness=-1

One of the significant challenges in our work was the discovery that batch-normalization was preventing our models from learning the problem. Batch Normalization has become a common tool in the deep learning literature for both faster convergence and a regularizing effect that often improves generalization~\cite{ioffe2015batch}. This makes the failure of batch-norm on our data an interesting and unique result, which we discuss in \autoref{sec:batchnorm_fail}. 

\subsection{On Failed Architectures} \label{sec:failed_models}

A large number of alternative architecture designs were tested for this problem, including up to 13 layers of convolution, using various (Bidirectional) RNNs, and with different attention models. The MalConv architecture presented performed best amongst many candidates. We review the other high level alternative architecture strategies here, the reasons why they failed to outperform our simpler MalConv, and how these relate back to out final design. Additional details can be found in the appendix.

Adding more layers is possible at the cost of decreased batch size, due to the aforementioned large memory use for backpropagation. We tested this with up to 13 layers of convolutions, and found performance only decreased. Many of these experiments  tried  smaller convolutional fields, so that the total receptive field of a neuron was on the scale of 500 to 1000 time steps. The problem with these approaches, beyond increasing training time to an untenable degree, is that it is not possible to due the standard approach of doubling the number of convolutional filters after each round of pooling to keep the amount of state per layer roughly equivalent. The state of the convolutions after 2 million steps is simply too large to reasonably compute on. Thus a rapid compression of state size per layer is necessary, but this ends up inhibiting learning. In our approach we have moved large amounts of information into the wide filter width in a single convolution, allowing us to exercise and retain information without exploding memory use. 

Another design choice was in processing the entire file simultaneously in one large convolution. An appealing notion would be to break up the input into chunks of 500 to 10,000 bytes, and process each chunk independently, as this would greatly reduce the training requirements. We tested this approach, and while it achieved reasonable  accuracies up to 95\%, it often failed to generalize to new data --- obtaining test accuracies in the 65-80\% range. This is because much of the contents of a given binary may be fully non-informative to a maliciousness decision, and training on random chunks and assuming a malicious label then encourages the model to overfit to the training data, and memorize the contents to produce correct decisions. Our MalConv model has access to the entire file which allows the model to detect the few informative features regardless of location. This is necessary to avoid the above variety of overfitting, and is objectively necessary to work in situations where normally benign programs have had malware injected into them. In this common situation most of the file should correctly indicate a benign program, while only a small fraction of the content is malicious.

The issue of information sparsity is also a factor in our choice to use temporal max-pooling rather than average-pooling. Beyond providing better interpretability, max-pooling also provided superior performance relative to average-pooling. The latter enforces a prior that informative features should be widely occurring in the underlying file. But many features will occur only once in the file, and so when combined with average-pooling, that feature's high response in one region of the binary will be washed-out by the remaining majority of the file that produces a low activation. Max-pooling avoids this problem, while still allowing us to tackle the variable-length issue. 

While RNNs are a common tool for any sequence related task, we found they reduced test accuracy when applied after our convolutions, by breaking the output after each convolution into a number of fixed sized chunks (with the last chunk containing padding). While an intuitive step to take, this also imposes a prior into the model that data coming from the convolution must regularly produce the same activation patterns at fixed frequencies. This is because the input to the RNN is re-shaping the temporal outputs of the CNN into a non-temporal matrix multiplication, and thus mandates the temporal information appear in consistent locations with a period equal to whatever chunk size was determined. This is not something the CNN can reasonable learn, and so performance is reduced. 

\section{Results}

We now present the results of our neural network model. To evaluate its performance and effectiveness, we will look at standard measures of accuracy in \autoref{sec:model_acc}, investigate the generalization capability of the learned features in \autoref{sec:manual_analysis}, and address batch-normalization issues in \autoref{sec:batchnorm_fail}. 
We will also take a moment to note the computational constraints required to build this model. To get the model to converge in a timely manner, we had to use a relatively larger batch size of 256 samples per batch.  Due to the extreme memory use of the architecture, this could not be performed on a single GPU. We were able to train this model on the 400k Group B set using data parallelism across the 8 GPUs of a DGX-1 in 16.75 hours per epoch, for 10 epochs, and using all available GPU memory. Training on the larger 2 million set took one month on the same system. 

\subsection{Malware classification} \label{sec:model_acc}

In evaluating the predictive performance of our models, we use Balanced Accuracy~\cite{Brodersen:2010:BAP:1904935.1905533} (i.e., accuracy weighted so that benign and malicious samples count evenly) and AUC~\cite{Bradley1997}
. We use balanced accuracy so that our results across the Group A and Group B tests sets are directly comparable, as they have differing proportions of benign and malicious samples. AUC is an especially pertinent metric due to the need to perform malware triage, where a queue of binaries to look at is created based on a priority structure~\cite{Jang2011}
. It is desirable to have the most malicious files ranked highest in the queue, so that they are identified and quarantined sooner rather than later. An analyst's time is expensive, and characterizing a single binary can take in excess of 10 hours~\cite{Mohaisen:2013:UZA:2487788.2488056}. A high AUC score corresponds to a successful ranking of most malware above most goodware, making it a directly applicable metric to evaluate. We pay particular attention to the accuracy on the Group A test set, as it has the fewest correlations with the Group B training set. Thus accuracy performance on Group A serves as a stronger measure of \textit{generalization} performance. In this vein we are also interested in which models have the smallest difference in performance between Groups A and B, which would indicate a model hasn't overfit to the source distribution. 

\begin{table*}[!htbp]
\centering
\caption{Performance of models on Group A and Group B test sets. Best results in \textbf{bold}, second best in \textit{italics}. }
\label{tbl:accuracy_results}
\begin{adjustbox}{max width=\textwidth}
\begin{tabular}{@{}lcccccccc@{}}
\toprule
         & \multicolumn{2}{c}{MalConv} & \multicolumn{2}{c}{MalConv w/o DeCov}  & \multicolumn{2}{c}{Byte n-grams} & \multicolumn{2}{c}{PE-Header Network}  \\ 
\cmidrule(lr){2-3} \cmidrule(lr){4-5} \cmidrule(lr){6-7} \cmidrule(lr){8-9}
Test Set & Accuracy       & AUC        & Accuracy            & AUC             & Accuracy          & AUC  & Accuracy          & AUC          \\
\midrule
Group A  & \textit{88.1}           & \textbf{98.5}       & 83.3                & \textit{98.4}           & 87.0              & \textit{98.4}  & \textbf{90.8} &   97.7    \\
Group B  & \textit{89.6}           & \textit{95.8}       & 86.6                & 94.3           & \textbf{92.5}              & \textbf{97.9}  & 83.7 &   91.4    \\ 
\bottomrule
\end{tabular}
\end{adjustbox}
\end{table*}

Despite the difficulty of the task at hand, we found that our networks tend to converge quickly, after only three epochs through the training corpus. This is in some ways beneficial, as the training time per epoch is significant. We believe this fast convergence may be due in part to the small size of our architecture, which has (only!) 134,632 trainable parameters. The accuracy results are shown in \autoref{tbl:accuracy_results}. Our model is able to achieve high AUCs when trained with and without regularization, indicating they would be useful for malware triage to help ranking of work queues.

Looking at the results, we can see our MalConv model is best or second best in performance on both metrics and test-sets. It also has the smallest performance difference between Group A and B test sets, indicating the model is using features that generalize well across the distributions. The byte n-gram model has high accuracy and AUC on the Group B test set, but the model also has a wide gap between Group A and B performance, indicating overfitting~\cite{raff_ngram_2016}. The byte n-gram model is also fragile to single-byte changes in the input, which will cause a feature to effectively "disappear" form the model's consideration. This is important when we consider that malware is written by an adversary capable of effecting such changes, making byte n-gramming a suboptimal approach. Our MalConv architecture does not have this same issue, and would require considerably more work to circumvent. Using a model trained on the PE-Header generalized well to the Group A test data, achieving a slightly higher accuracy than MalConv, but has significantly reduced performance on Group B in terms of accuracy and AUC. This shows some robustness, but indicates the same features aren't being used equally across domains. Overall, MalConv provides the most encouraging balance in performance across all data and metrics. 

The application of DeCov regularization significantly improves the accuracy of the model for both Group A and B test sets. This is a somewhat unusual property, as it appears that the DeCov's primary impact is to improve the calibration of the decision threshold, rather than the underlying concept learned by the model. This was a problem noted in \textcite{raff2017peheader} for their PE-header network. 
Applying DeCov has successfully improved the calibration of the model's output probabilities, increasing the accuracy by up to 4.8 points. 

\begin{table}[!tb]
\centering
\caption{Performance of models on Group A and Group B test sets, when using new 2 million training corpus. Best results in \textbf{bold}}
\label{tbl:results_2m}
\begin{tabular}{@{}lcccc@{}}
\toprule
\multicolumn{1}{c}{} & \multicolumn{2}{c}{MalConv}   & \multicolumn{2}{c}{Byte n-grams} \\ 
\cmidrule(lr){2-3} \cmidrule(lr){4-5}
Test Set             & Accuracy      & AUC           & Accuracy            & AUC        \\ \midrule

Group A              & \textbf{94.0} & \textbf{98.1} & 82.6                & 93.4       \\
Group B              & 90.9          & \textbf{98.2} & \textbf{91.6}       & 97.0       \\ \bottomrule
\end{tabular}
\end{table}

Using a larger corpus of 2 million files, we can also see that the MalConv model improves its performance, increasing Group A and B accuracy by 5.9 and 1.3 points, and Group B AUC by 2.4 points. We have also replicated the byte n-gram model that the original Group B training data used, and found that performance dropped on the Group A test set by 4.4 points for accuracy and 5.0 points for AUC. Group B test performance was also reduced, though not significantly. This highlights the predicted brittleness and propensity for overfitting of byte n-grams for malware detection \cite{raff_ngram_2016}. Our MalConv network's improvement with more data highlights its superiority, and that it has greater capacity to tackle this problem than prior domain knowledge free approaches.

\subsection{Manual Analysis} \label{sec:manual_analysis}

Using our architecture design, we are able to perform a modest manual analysis of what the model has learned. We do this by adapting the approached used by \textcite{Zhou2016}, which produces a \textit{class activation map} (CAM) for each class in the output. We use a global max-pooling layer in our work, rather than the average pooling originally proposed. Doing so produces a naturally sparse activation map which aids interpretability, which we call a sparse-CAM. This is a critical design choice given the extreme sequence length of our binaries, as it would be impractical to examine all 2 million bytes. This sparse-CAM design will return one 500 byte region as "important" for each convolutional filter; since our model uses 128 filters, there are at most 128 regions marked for each binary.

This approach enables us to produce CAM mappings for regions that are indicative of benignness or maliciousness to the learned network. This preference towards benign or malicious is determined by the sign of the produced activation map. Using the PE-file library~\cite{pefileGithub}, we can parse most of our binaries into different regions. These regions correspond to different portions of the binary format. For example, there is a PE-Header that specifies the regions of the file. We expect any approach to learn significant information from this region, as it is the most structured and accessible portion of a binary. The PE-Header then identifies which sections of the binary store the executable code (\textit{.text} or \textit{CODE} sections), global variables (\textit{.data}), and others. By determining which region each sparse-CAM occurred in, we can gain insights about what our model is learning. We show the results of this applied to 224 (7 mini-batches) randomly selected binaries from the Group A test set. This allows us to best evaluate the generalized knowledge of the network, and the results are shown in \autoref{tbl:sec_important_features}. 

\begin{table*}[!htbp]
\centering
\caption{Important features as determined by section, as determined by the non-zero regions of the sparse-CAM mapped to the output of PE-file.}
\label{tbl:sec_important_features}
\begin{tabular}{@{}lccccccccc@{}}
\toprule
Section   & Total  & PE-Header & .rsrc & .text & UPX1 & CODE & .data & .rdata & .reloc \\ \midrule
Malicious & 26,232 & 15,871    & 3,315 & 2,878 & 697  & 615  & 669   & 383    & 214    \\
Benign    & 19,290 & 11,183    & 2,653 & 2,414 & 596  & 505  & 423   & 243    & 77     \\ \bottomrule
\end{tabular}
\end{table*}

Previous work building byte n-gram models on this data found that byte n-gram's obtained almost all information from the PE-Header~\cite{raff_ngram_2016}. 
Based on the sparse-CAM locations, we find that only 58-61\% of information MalConv is using also comes from the PE-Header, indicating a larger diversity of information types are being used. The \textit{.rsrc} section indicates use of the resource directory, where contents like file icons (but also executable code) may be stored. Importantly we also see the \textit{.text} and \textit{CODE} sections activating, indicating that our model is using some amount of executable code as a feature. Similarly, application data found in \textit{.data} and \textit{.rdata} indicates our model may be detecting common structural patterns between binaries. 

We note in particular that the \textit{UPX1} section has been indicative of both benign and malicious binaries, as learned by our network. The \textit{UPX1} section indicates the use of packing, specifically the widely used UPX packer~\cite{upxGithub}. Packing will compress or encrypt most of the binary into a single archive which is extracted at runtime. This makes simple static analysis difficult, and packing is prevalent among malware authors to hinder malware analysis. However, packing alone is not a reliable malware indicator, as many benign applications are also packed~\cite{Guo:2008:SPP:1433006.1433014}. The prevalence of packing in malicious executables leads to many models learning a direct (but unhelpful) equivalence between "packed" and "malicious". Our results indicate that our model may have avoided such an association. We hope further advances in interpretable models will help us to confirm this behavior, and determine which minute details allow the model to change its inclination.

\subsection{The Failure of Batch-Normalization} \label{sec:batchnorm_fail}

Our results are seemingly in conflict with what has been reported in numerous other works, since the addition of batch-normalization to MalConv consistently failed to learn after several epochs.
At best models trained with batch-norm would obtain 60\% training and 50\% test accuracies. This phenomena occurred with all architecture design variants. 
Our surprise at this result lead us to implement this, and other, architectures using batch-normalization in PyTorch, Tensorflow, Chainer, and Theano. Batch-norm failed to converge or generalize in all cases. 

\begin{figure}[!htb]
\centering
\begin{tikzpicture}[scale=1.0]
\begin{axis}[
  height=0.35\textwidth,
  width=1.0\columnwidth,
  xmin=-4,
  xmax=4,
  xlabel=Standardized Output Value,
  ylabel=PDF,
  y label style={at={(-0.1,0.5)}},
  legend pos=north east]
\addplot +[mark=none, dash dot,red,thick] table [y=pdf, x=x,col sep=comma]{data/kde_res5c_branch2c.csv};
\addlegendentry{Res5c}
\addplot +[mark=none, dash dot,orange,thick] table [y=pdf, x=x,col sep=comma]{data/kde_res3b3_branch2c.csv};
\addlegendentry{Res3b3}
\addplot +[mark=none, dash dot,green,thick] table [y=pdf, x=x,col sep=comma]{data/kde_conv1_7x7_s2.csv};
\addlegendentry{Iv4 Conv1}
\addplot +[mark=none, blue,thick] table [y=pdf, x=x,col sep=comma]{data/kde_mal_decov.csv};
\addlegendentry{MalConv}
\addplot +[mark=none, black,dashed,thick,samples=50] {gauss(0, 1.0)};
\addlegendentry{ $\mathcal{N}(0,1)$}

\end{axis}
\end{tikzpicture}
\caption{KDE plots of the convolution response (pre-ReLU) for multiple architectures. Red and orange: two layers of ResNet green: Inception-v4 blue: our network; black dashed: a true Gaussian distribution for reference.}
\label{fig:kde_conv_responses}
\end{figure}
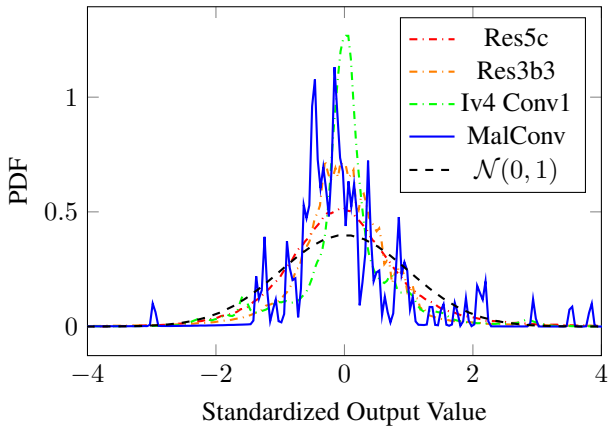

To diagnose this problem, we started with the fact that batch-normalization assumes that data should be re-fit to a unit-normal distribution. We then plotted the pre-activation function response of layers in our network along with that of the Gaussian distribution, which can be seen in \autoref{fig:kde_conv_responses}. The figure shows kernel density estimates of the responses from earlier layers in networks trained on images or on binary executables. Networks trained on image data display an approximately Gaussian distribution of activations (smooth and unimodal), while the activation distribution of our network exhibits much greater 
asperity. %
Since batch normalization assumes the data to be normalized is normally distributed, this may account for its ineffectiveness in our application.
We recommend that any applications of batch-normalization to new problems produce similar such visualizations as a method to diagnose convergence issues. 

\pgfmathdeclarefunction{gauss}{2}{%
  \pgfmathparse{1/(#2*sqrt(2*pi))*exp(-((x-#1)^2)/(2*#2^2))}%
}

We hypothesize that batch norm's ineffectiveness in our model is a product of training on binary executables. The majority of contemporary deep learning research, including batch-normalization, has been done in the image and signal processing domains, with natural language a close second. In all of these domains the nature of data is relatively consistent. In contrast, our binary data presents a novel multi-modal nature of the byte values. The same byte value can have drastically different meaning depending on the location, ranging from ASCII text, code, structured data, or even images stored for the icon. Our hypothesis is that this multi-modal nature produces multiple modes of activation, which then violates the primary assumptions of batch-normalization, causing degraded performance. 

Our tests in using models trained on random chunks of only 500 to 10,000 bytes of the binary support this hypothesis. When trained on a random sub-region like this, the majority of bytes will be of a single modality when presented, and thus present a smoother unimodal activation pattern. This was the only case where batch-norm was able to reach high training accuracies above 60\% for our data, but still did not generalize to the test data (obtaining only 50\% random-guessing accuracy). 

\section{Conclusion}

In this work we have described the use of neural networks on the raw bytes of entire executable files. This solution avoids a number of the issues with the more common byte n-gram approach, such as brittle features and over-focusing on the PE-Header as important information. It achieves consistent generalization across both test sets, despite the challenges of learning a sequence problem of unprecedented length.

In a broader machine learning context, we have identified a number of unique learning challenges and discussed techniques for addressing classification of extremely long sequences. Our work has extended the application of neural networks to a domain beyond images, speech, etc. to one with much more sophisticated spatial correlation behaviors. In doing so, we identify a potential pitfall with the very commonly used batch-normalization and suggest a way to check if the technique is appropriate (a normality test of pre-activation function response). 

In future work we hope to further developed architectures that work in this domain, 
to further explore the batch-normalization issue, and determine what types of existing normalization or weight initialization schemes work with such multi-modal responses. Critical thought must also be given to ways in which the memory intensive nature of this problem can be reduced, and what types of architectural designs may allow us to better capture the multiple modes of information represented in a binary.  A general approach to byte level understanding of programs would have many applications beyond malware classification such as static performance prediction and automated code generation.

\subsubsection*{Acknowledgments}

Special thanks to Mark McLean of the Laboratory for Physical Sciences for supporting this work.

\clearpage

\FloatBarrier
\section*{Appendix}

\section{Discussion of Alternative Architectures}

We have exhausted many GPU compute years experimenting with 
various alternative neural network architectures for this problem, none of which performed as well as the MalConv approach presented here.  It is not feasible to provide a complete discussion or numerical comparison of architectures tried, but we will give an overview of our experience with some significant design choices. This section expounds upon additional details and experiments that could not be included in \autoref{sec:failed_models} due to space constraints. Additional references are provided to techniques we discuss in this section, but are not relevant to the discussion of the original paper. 

\subsection{Architectural Approaches}

As mentioned in our primary paper, memory constraints are the primary bottleneck we must work around to build a model on such long time series. One straightforward notion to tackle this problem would be to attempt to train on separate smaller sub-regions of the input independently.  We tried this with regions of width 500, 1000 and 10,000 and discovered with this approach that the majority of the binary may be non-discriminative for this problem because it contains standard or common code that any application (benign or malicious) would need. As such the labels for sub-sections were overwhelmingly noisy as they had to be derived from the global label for the whole input sequence.  These models overfit strongly to the training data and failed to generalize to the test data. While applying multiple strong regularizes could get such models to generalize to some degree (test accuracies in the 80\% range), they still underperformed compared to MalConv. In addition, such models had high variability in training --- and on average only one in ten models would converge to something usable, where most would degrade to random guessing. We conclude that processing the whole sequence at once is important to ensure that whatever smaller features occur that are discriminative can be extracted regardless of location in the sequence.  As such we focused our efforts on shallower architectures with smaller numbers of filters in the early layers.

This result was supported by another test, where we took models trained on random regions of the binary, and at inference time, applied them to multiple random regions and averaged the results. One might hope this would have the effect of creating an ensemble of models, but in practice had no impact on the accuracy of our models. This test helped us to conclude on the importance of having the entire scope of the binary being a necessary component of our model.

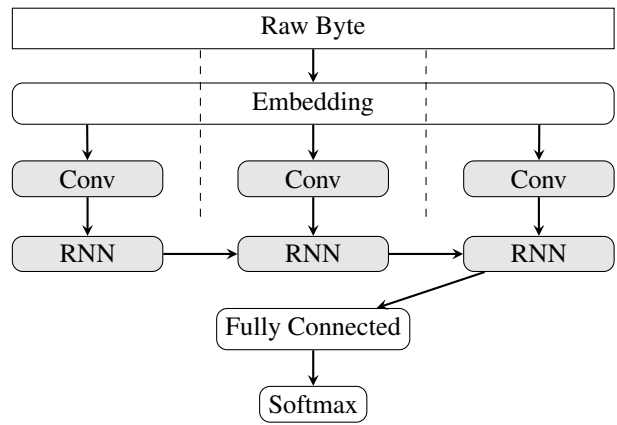
\begin{figure}[!htb]
\begin{adjustbox}{max width=\columnwidth}
\begin{tikzpicture}[auto]

      \node (byte) [rectangle, draw=black,minimum width=8cm] {Raw Byte};

      \node (embd) [rectangle, rounded corners, draw=black, below of=byte,minimum width=8cm] {Embedding};

      \node (conv_1) [rectangle, rounded corners, draw=black, below of=embd, minimum width=2cm,fill=gray!20] {Conv};
      
      \node (conv_2) [rectangle, rounded corners, draw=black, left of=conv_1, minimum width=2cm,xshift=-2.0cm,fill=gray!20] {Conv};
      
      \node (conv_3) [rectangle, rounded corners, draw=black, right of=conv_1, minimum width=2cm, xshift=+2.0cm,fill=gray!20] {Conv};
      
      \node (rnn_1) [rectangle, rounded corners, draw=black, below of=conv_1, minimum width=2cm,fill=gray!20] {RNN};
      \node (rnn_2) [rectangle, rounded corners, draw=black, below of=conv_2, minimum width=2cm,fill=gray!20] {RNN};
      \node (rnn_3) [rectangle, rounded corners, draw=black, below of=conv_3, minimum width=2cm,fill=gray!20] {RNN};
      
      \node (flc) [rectangle, rounded corners, draw=black, below of=rnn_1] {Fully Connected};

      \node (out) [rectangle, rounded corners, draw=black, below of=flc] {Softmax};

\draw [thick,->,>=stealth] (byte) --  node[anchor=east] {} (embd);

\draw [thick,->,>=stealth] (embd) --  (conv_1);
\draw [thick,->,>=stealth] ($(embd.south)!0.75!(embd.south west)$) --  (conv_2);
\draw [thick,->,>=stealth] ($(embd.south)!0.75!(embd.south east)$) --  (conv_3);
\draw[dashed] let \p1=(conv_2),\p2=(conv_1),\p3=(conv_1.north west),\p4=(conv_1.south west) in 
	({(\x1+\x2)/2},\y3-0.75cm) -- ({(\x1+\x2)/2},\y4+2cm);
\draw[dashed] let \p1=(conv_3),\p2=(conv_1),\p3=(conv_1.north west),\p4=(conv_1.south west) in 
	({(\x1+\x2)/2},\y3-0.75cm) -- ({(\x1+\x2)/2},\y4+2cm);
    
\draw [thick,->,>=stealth] (conv_1) --  (rnn_1);
\draw [thick,->,>=stealth] (rnn_2) --  (rnn_1);
\draw [thick,->,>=stealth] (conv_2) --  (rnn_2);
\draw [thick,->,>=stealth] (rnn_1) --  (rnn_3);
\draw [thick,->,>=stealth] (conv_3) --  (rnn_3);

\draw [thick,->,>=stealth] (rnn_3) --  (flc);

\draw [thick,->,>=stealth] (flc) --  (out);

\end{tikzpicture}
\end{adjustbox}
\caption{"Chunk" strategy where the embedding is split into parts. Dashed lines indicate outputs that were split into parts. Boxes with gray backgrounds and the same name indicate weight sharing. } \label{fig:chunk_break_embed}
\end{figure}

In \autoref{sec:failed_models}, we also discussed breaking up the file into chunks to be used in a RNN. We clarify the details of these attempts here. The first attempted strategy was to break the embedding into $N$ separate adjacent portions after the embedding, which are then passed to the same convolution filters (see \autoref{fig:chunk_break_embed}). The figure depicts a simple RNN architecture, but this chunking strategy was tried with various combinations of techniques. In addition, each layer type in the diagram was tested for multiple depths and sizes. The figure is only to illustrate the general high level architecture form. 

Breaking up the embedding into contiguous chunks was done in an attempt to exploit model parallelism in Tensorflow, so that larger sequences of convolutions and filters could be attempted with each chunk being processed by a single GPU, thus giving them more memory to support such work.

The above chunking strategy did not perform well. One issue encountered was that breaking the chunks into many smaller pieces caused artifacts in the output near the boarders of each chunk. This was a potential problem in part because most of our architectures had a receptive field of 500 to 1000 time steps, meaning a non-trivial portion of the chunked convolution outputs would have 
significant edge effects at the borders of each chunk. Since this strategy inherently produces many chunks, the edge effect is multiplied --- thus negatively impacting performance.

\begin{figure}[!htb]
\begin{adjustbox}{max width=\columnwidth}
\begin{tikzpicture}[auto]

      \node (byte) [rectangle, draw=black,minimum width=8cm] {Raw Byte};

      \node (embd) [rectangle, rounded corners, draw=black, below of=byte,minimum width=8cm] {Embedding};

      \node (conv_1) [rectangle, rounded corners, draw=black, below of=embd, minimum width=8cm] {Conv};

      \node (rnn_1) [rectangle, rounded corners, draw=black, below of=conv_1, minimum width=2cm,fill=gray!20] {RNN};
      \node (rnn_2) [rectangle, rounded corners, draw=black, left of=rnn_1, xshift=-2.0cm, minimum width=2cm,fill=gray!20] {RNN};
      \node (rnn_3) [rectangle, rounded corners, draw=black, right of=rnn_1, xshift=+2.0cm, minimum width=2cm,fill=gray!20] {RNN};
      
      \node (flc) [rectangle, rounded corners, draw=black, below of=rnn_1] {Fully Connected};

      \node (out) [rectangle, rounded corners, draw=black, below of=flc] {Softmax};

\draw [thick,->,>=stealth] (byte) --  node[anchor=east] {} (embd);

\draw [thick,->,>=stealth] (embd) --  (conv_1);
\draw [thick,->,>=stealth] (conv_1) --  (rnn_1);
\draw [thick,->,>=stealth] ($(conv_1.south)!0.75!(conv_1.south west)$) --  (rnn_2);
\draw [thick,->,>=stealth] ($(conv_1.south)!0.75!(conv_1.south east)$) --  (rnn_3);
\draw[dashed] let \p1=(rnn_2),\p2=(rnn_1),\p3=(rnn_1.north west),\p4=(rnn_1.south west) in 
	({(\x1+\x2)/2},\y3-0.00cm) -- ({(\x1+\x2)/2},\y4+1.85cm);
\draw[dashed] let \p1=(rnn_3),\p2=(rnn_1),\p3=(rnn_1.north west),\p4=(rnn_1.south west) in 
	({(\x1+\x2)/2},\y3-0.0cm) -- ({(\x1+\x2)/2},\y4+1.85cm);
    
\draw [thick,->,>=stealth] (rnn_2) --  (rnn_1);
\draw [thick,->,>=stealth] (rnn_1) --  (rnn_3);
\draw [thick,->,>=stealth] (rnn_3) --  (flc);

\draw [thick,->,>=stealth] (flc) --  (out);

\end{tikzpicture}
\end{adjustbox}
\caption{"Chunk" strategy where the convolution is split into parts. Dashed lines indicate outputs that were split into parts. Boxes with gray backgrounds and the same name indicate weight sharing. } \label{fig:chunk_break_conv}
\end{figure}
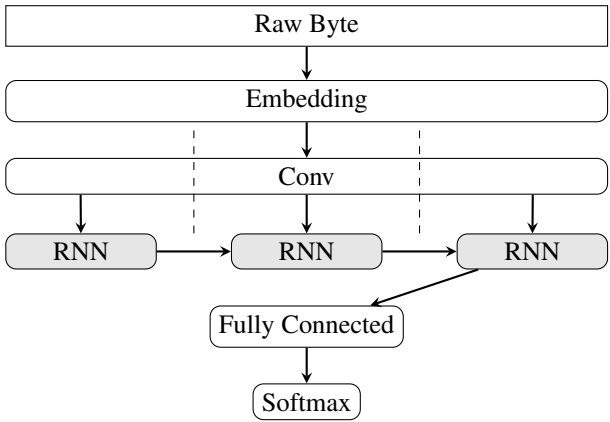

To test how large an impact this had, we moved the "chunking" past the embeddings and after the convolutions, as shown in \autoref{fig:chunk_break_conv}. This is somewhat similar to early work in object detection with CNNs \cite{girshick2015fast}. We also hoped that this would improve throughput with respect to the first chunking strategy, giving a chance for the GPU kernels to operate on more data at a time. This performance improvement did not materialize in a meaningful way, and came at the cost of forcing the use of only data-parallelism.

Ultimately these chunking strategies did not perform as well as MalConv, regardless of the method used to combine the chunks in the end. LSTMs and GRUs were tested to take the output, using unidirectional and bidirectional variants. In addition, each variant was tested using both the last RNN's hidden state to feed into the final fully connected layers, simple averaging of the states produced after each timestep, and the attention mechanism proposed in \citet{raff2017peheader}.

\subsubsection{Malware "Images"}

We make note of a particular architectural strategy that is referred to as a malware "image" \cite{Nataraj:2011:MIV:2016904.2016908}. Such images are generally constructed by treating each byte of the binary as a gray-scale pixel value, and defining an arbitrary "image width" that is used for all images. This technique was developed to try and exploit early image-processing approaches as a way of visualization, and then for malware family classification \cite{Nataraj:2011:CAM:2046684.2046689}. This strategy may at first seem intuitive as it would allow one to leverage existing image-classification techniques, but it results in a number of problems. 

The problems that arise stem from the fact that binaries are not images, and construing them as such introduces priors into the model that are objectively false. First, we note that creating images in this fashion necessitates the selection of an image width, which becomes a new hyper-parameter in the problem that must be tuned. Selecting the width then determines the height based on the number of rows in the image, for which the last row many not be complete -- and some strategy will be needed to impute the last pixels. 

This then leads to the question, how does one handle the variable number of rows each image will have? The aforementioned work by \textcite{Anderson2017} chose the option of truncating all binaries to only 256 KB, which only partially resolves the image height problem. This is because there may be binaries smaller than 256 KB, and it also means the approach can not process all the raw bytes of an executable. From an adversarial perspective, the approach can be trivially circumvented by any malware which makes itself large enough and inserts it's malicious payload into the end of the file. Other approaches to dealing with variable height are equally unsatisfying. For example, one can not meaningfully re-scale a malware image like one can for a normal image. Doing so would interpolate pixel values which correspond to exact byte values, acting on an interpretation that byte values are numerically meaningful, which is objectively false. 

\begin{figure}[!htb]
  \begin{center}
    \includegraphics[width=1.0\columnwidth]{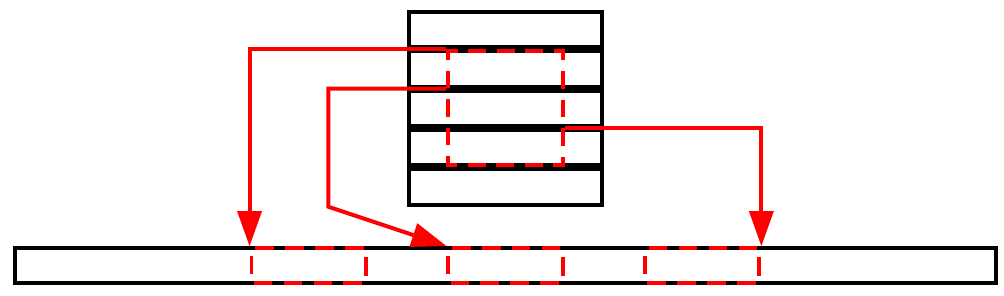}
  \end{center}
  \caption{Simple demonstration of the spatio-temporal problem caused by creating malware "images". The red dashed area shows the receptive field of a convolution, mapped from the malware image form (top) back to the raw byte sequence (bottom). } 
  \label{fig:mal_img_probl}
\end{figure}

In the context of using modern deep learning architectures developed for tasks like image-net, we note that the 2D convolution of the malware image imposes non-existent gaped spatial correlations. A visualization of this is given in \autoref{fig:mal_img_probl}, where a 2D convolution on a malware image is unwrapped to correspond to the original byte-sequence. In this context is becomes clear that a 2D convolution of a malware image is akin to a dilated 1D convolution, where the gaps in the convolution are determined by the image width, and each filter processes a contiguous group of "pixels" at a time. As discussed in the main paper, dilated convolutions were tested for the MalConv architecture, but had worse performance - likely because the gaps in the dilation are not meaningfully inferable like they are for natural images which have strong spatial consistency.  While the given figure shows small gaps for brevity, we note in practice the gaps will be large. This imposes a false prior into the model that bytes are correlated in fixed patterns across large ranges of the binary, which does not make sense. 

For these reasons we do not consider the malware image approach in this work, and that it is unlikely to produce satisfying solutions. Indeed, as we have shown, it is little more than a dilated 1D convolution. 

\subsection{Architecture Parameters}

Initially we attempted to use deeper networks (up to 13 layers) with narrower convolutional filters (width 3--10), smaller strides (1--10) and organized into residual blocks. This was largely motivated by the trends in current deep learning literature, which has widely converged on smaller filter sizes stacked depth-wise to produce larger receptive fields and representative power \cite{he15deepresidual,NIPS2015_5850}. This did not translate to our new domain, where we found reversing both trends (fewer layers and wider filters) gave us our initial successes. In order to facilitate tests of deeper networks in our architecture, we also attempted freezing the weights of certain layers so that they would not be trained, thus reducing the amount of activations that need to be kept to perform back-propagation, and thus, reduce memory usage. We attempted freezing just the first few convolutions, every $n$ convolution, and even having a network of purely random convolutions. While able to achieve high training accuracy, most of these random convolutions had sub-par test performance (60\% or less). While it is not clear that a shallow approach is intrinsically superior for our problem, it did outperform the myriad deeper approaches we implemented.

We desired testing many more architectures, but from a practical standpoint those which do not perform rapid spatial compression are currently beyond our (not insignificant) compute capabilities. The large memory footprint of the activations from the first convolution layers in these cases forced us to use very small effective batch sizes (4--8 per GPU) and resulted in impractically slow training without obvious convergence after weeks-to-months of runtime. 

Within our experiments with convolutional architectures we also tried varying the number of filters, but more filters often led to greater overfitting.  In a similar trend, increasing the size of the embedding space also lead to increased overfitting. We tried dilated convolutions to see if a larger receptive field was helpful but, again, struggled with overfitting.  Un-gated convolutions led to similar but inferior performance compared to the MalConv architecture.

We also attempted to replace the convolutional layer in the MalConv architecture with a quasi-recurrent layer ~\cite{DBLP:journals/corr/BradburyMXS16}.  Whilst these networks converged they also overfit and achieved worse validation loss than MalConv.

Some other model regularization variants we tested that converged but led to worse final validation loss were: temporal average pooling instead of temporal max-pooling, weightnorm \cite{NIPS2016_6114} instead of DeCov regularization, Elastic-net regularization, regional dropout on the input, and gradient noise \cite{Neelakantan2016}. For experiments that used RNNs, we attempted both normal dropout \cite{Srivastava2014} and more modern Bayesian dropout \cite{NIPS2016_6241}. 

Our final MalConv architecture used the common ReLU activation function. We did tests on other activations such as ELU \cite{Clevert2016}, Leaky ReLU \cite{Maas2013}, and PReLU \cite{He:2015:DDR:2919332.2919814}. While not detrimental, we found no positive impact from their inclusion. 

For training the MalConv model, we found SGD with Nestrov momentum \cite{Bengio2013} worked best. The initial learning rate was set to $0.01$, the momentum term  set to $0.9$, and an exponential decay rate was used. Other update schemes such as Adam \cite{Kingma2015}, AdaDelta \cite{Zeiler2012}, and RMSProp \cite{Tieleman2012}, were all tested. For all update schemes multiple learning rates and decay rates were tested, congruent with the recommended ranges for each approach.

\begin{figure*}[!h]
  \begin{center}
    \includegraphics[width=0.8\textwidth]{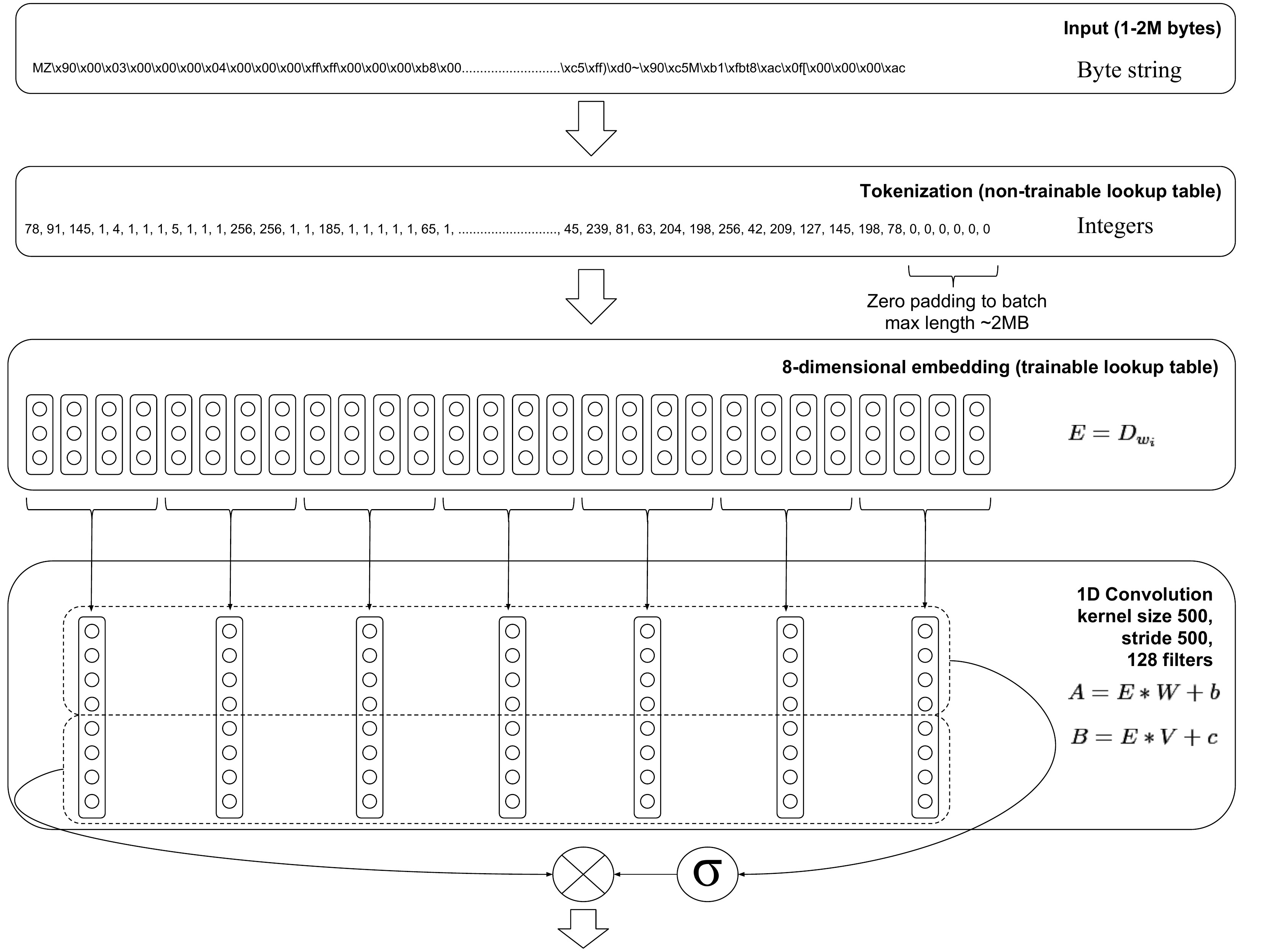}
    \includegraphics[width=0.8\textwidth]{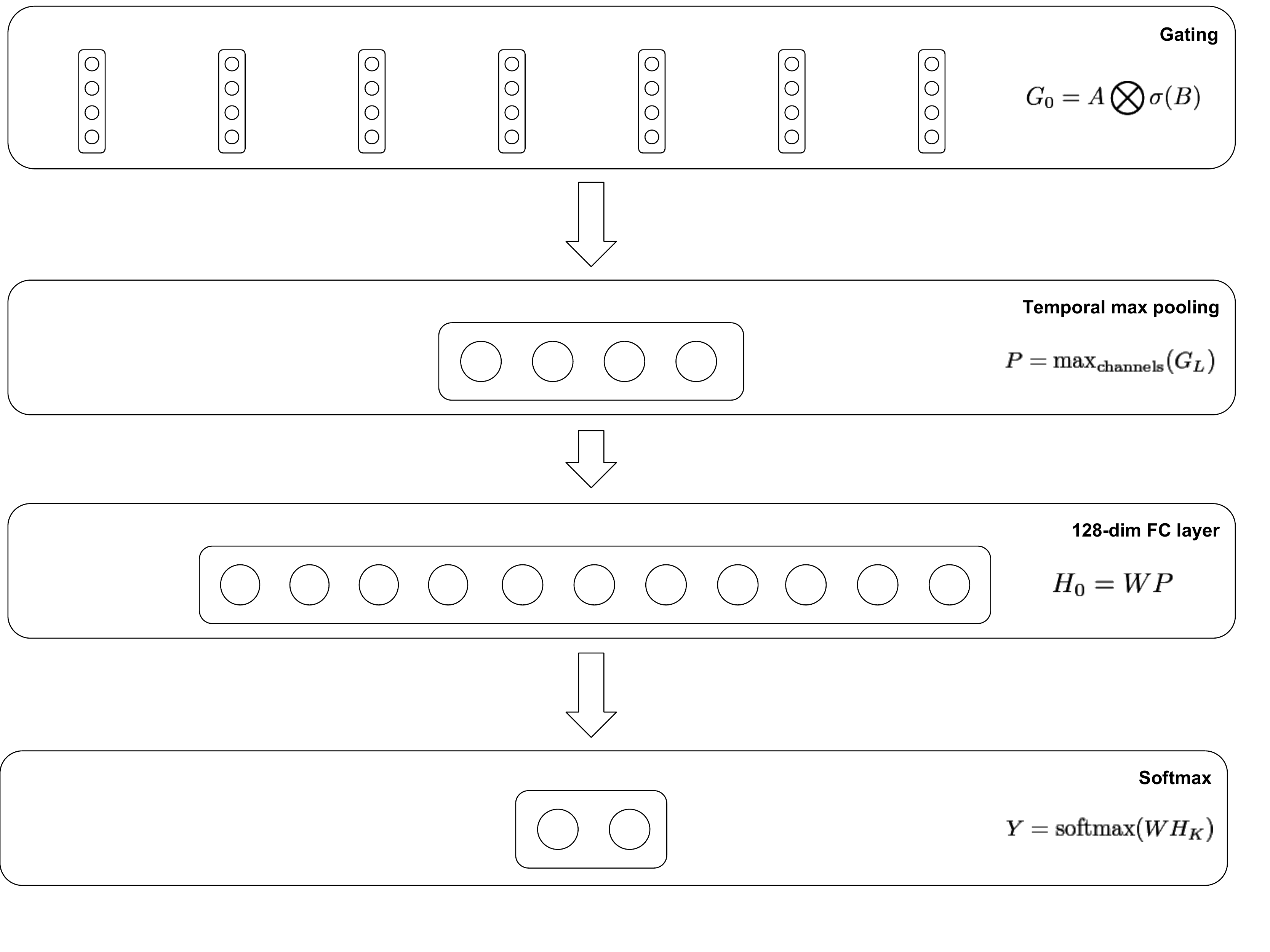}
  \end{center}
  \caption{Full architecture diagram of MalConv model.} 
  \label{fig:full_diagram}
\end{figure*}

\FloatBarrier

{\small
\bibliographystyle{aaai}
\bibliography{extracted,aaai_appendix}
}

\end{document}